\documentclass[runningheads]{llncs}
\usepackage{graphicx}
\usepackage{hyperref}

\usepackage{amsmath,amssymb} 
\usepackage{color}
\usepackage{subfig}
\usepackage{booktabs}
\usepackage{multirow} 
\usepackage{enumitem}
\usepackage{wrapfig}

\definecolor{darkgreen}{RGB}{0,160,0}
\definecolor{orange}{RGB}{238,118,0}

\newcommand\vs{\textit{vs.}}
\newcommand\eg{\textit{e.g.}}
\newcommand\ie{\textit{i.e.}}

\usepackage{floatrow}
\newfloatcommand{capbtabbox}{table}[][\FBwidth]

\begin{document}
%
\title{CLIP Meets Video Captioning: Concept-Aware Representation Learning Does Matter}

\titlerunning{CLIP Meets Video Captioning}
%

\author{}
\authorrunning{}
\institute{}

\author{Bang Yang\inst{1,2} \and
Tong Zhang\inst{2} \and
Yuexian Zou\inst{1,2}\thanks{Corresponding author.}}
\authorrunning{B. Yang, T. Zhang and Y. Zou}
%
\institute{ADSPLAB, Shenzhen Graduate School, Peking University, Shenzhen, China \and
Peng Cheng Laboratory, Shenzhen, China \\
\email{\{yangbang,zouyx\}@pku.edu.cn; \{zhangt02\}@pcl.ac.cn}}

\maketitle              
\begin{abstract}
For video captioning, ``pre-training and fine-tuning'' has become a de facto paradigm, where ImageNet Pre-training (INP) is usually used to encode the video content, then a task-oriented network is fine-tuned from scratch to cope with caption generation. This paper first investigates the impact of the recently proposed CLIP (Contrastive Language-Image Pre-training) on video captioning. Through the empirical study on INP \vs{} CLIP, we identify the potential deficiencies of INP and explore the key factors for accurate description generation. The results show that the INP-based model is tricky to capture concepts' semantics and sensitive to irrelevant background information. By contrast, the CLIP-based model significantly improves the caption quality and highlights the importance of concept-aware representation learning. With these findings, we propose Dual Concept Detection (DCD) further to inject concept knowledge into the model during training. DCD is an auxiliary task that requires a caption model to learn the correspondence between video content and concepts and the co-occurrence relations between concepts. Experiments on MSR-VTT and VATEX demonstrate the effectiveness of DCD, and the visualization results further reveal the necessity of learning concept-aware representations.
\end{abstract}
\section{Introduction}
\label{sec:introduction}

Video captioning aims to describe video content with fluent sentences. Given the difficulties of learning effective video representations from limited data\cite{xu2016msr,wang2019vatex}, mainstream video captioning methods adopt the Encoder-Decoder framework \cite{venugopalan2015translating} with a ``pre-training and fine-tuning'' paradigm, where ImageNet Pre-training (INP) is usually used to help encode the video content, and a task-oriented network is fine-tuned from scratch to cope with caption generation. 
However, using INP across discrepant tasks may bring limited benefit \cite{he2019rethinking}.

\begin{figure}[t]
\centering
\includegraphics[width = 0.9\linewidth]{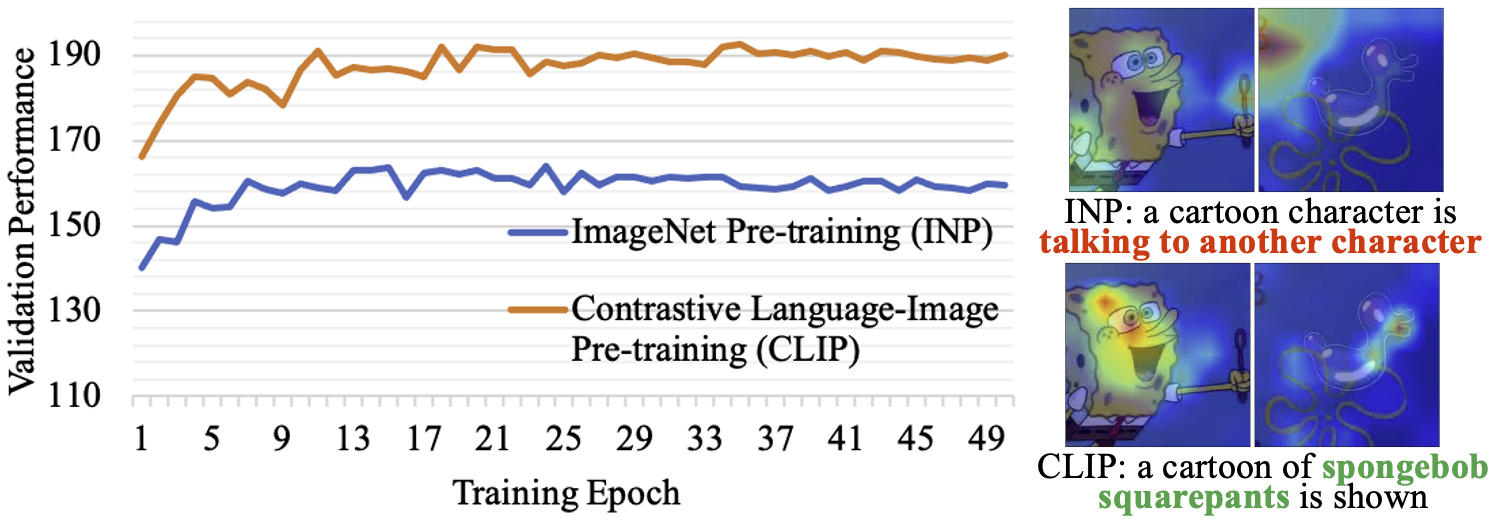}
\caption{ImageNet Pre-training (INP) \vs{} Contrastive Language-Image Pre-training (CLIP). When using CLIP rather than INP to help encode the video content, performance is greatly improved (left), which can be attributed to the better learning of concept-aware representations (right). The right part visualizes Grad-CAM \cite{selvaraju2017grad} for the ground-truth caption ``spongebob squarepants blows bubbles''. The caption below each example is model's actual prediction.
\setlength{\belowcaptionskip}{100pt}
}
\label{fig:INP_vs_VLP}
\end{figure}

Recent advances in video captioning \cite{chen2021motion,lin2021augmented,zhang2021open,zhang2020object,yang2021visual,liu2021o2na,li2022adaptive} are built upon the default use of INP and meanwhile, their performance gradually becomes saturated on MSR-VTT\cite{xu2016msr}. 
Thus, immediate questions raise: \textit{is INP causing a performance bottleneck for video captioning? If so, why?} To answer these questions, we turn our attention to CLIP (Contrastive Language-Image Pre-training) \cite{DBLP:conf/icml/RadfordKHRGASAM21}, which has drawn great attention in the community due to its strong zero-shot transfer ability to various vision tasks. 
As a branch of vision-language pre-training research \cite{sun2019videobert,li2020hero}, CLIP is unbounded by a fixed set of labels and its pre-training data, \ie{}, 400M noisy image-text pairs crawled from the Internet, is larger than ImageNet by an order of magnitude. 
To this end, we hypothesize that CLIP has great potentials for video captioning.

In this paper, we carry out an empirical study on INP \vs{} CLIP to shed new light on potential deficiencies of INP for caption generation and explore the key to prompting accurate video captioning. Fig.~\ref{fig:INP_vs_VLP} gives a snapshot of our experimental study. 
We can see from the curves that captioning performance is significantly improve when using CLIP rather than INP to help encode the video content. 
This performance gap can be interpreted by the visualized example shown on the right, where INP deviates the video caption model's focus from the critical regions of concepts ``spongebob squarepants'' and ``bubbles'' while CLIP's results just the opposite. 
As a result, the empirical study has shown that concept-aware representation learning does matter to accurate caption generation.

Based on the above finding, we propose Dual Concept Detection (DCD) to spur video caption models to learn concept-aware video and text representations during training. Specifically, DCD requires the the caption model to infer relevant concepts based on partial information from no matter videos or text descriptions. 
To achieve that, the model has to build the correspondence between video content and concepts and learn the concepts' co-occurrence relations.

To summarize, we make the following contributions. (1) We carry out an empirical study on INP \vs{} CLIP for video captioning. The results reveal the deficiencies of INP and suggest the importance of concept-aware representation learning in prompting accurate captioning. (2) Motivated by the success of CLIP for video captioning, we introduce Dual Concept Detection, an auxiliary task that can be jointly trained with video caption models to strengthen their learning of concept-aware representations during training. (3) Experiments on MSR-VTT\cite{xu2016msr} and VATEX \cite{wang2019vatex} verify the effectiveness of our approach and ablation studies clearly show what leads to the improvements.

\section{Related work}

\textbf{Video Captioning.} The ``pre-training and fine-tuning'' paradigm is commonly used from the very first neural-network-based methods \cite{venugopalan2015sequence,venugopalan2015translating} to the present day. Recent focuses in video captioning include but not limited to (1) learning fine-grained representation via graph neural models \cite{pan2020spatio,zhang2020object,chen2021motion} or hierarchical encoders \cite{yang2021visual}, (2) distilling knowledge mutually \cite{lin2021augmented} or from external models \cite{zhang2020object}, and (3) introducing new paradigms like non-autoregressive captioning \cite{yang2021non,liu2021o2na} and open-book captioning \cite{zhang2021open}. However, recent advanced methods pay less attention to the effect of pre-training models, suffering from potential performance bottleneck.

\noindent\textbf{Vision-Language Pre-training (VLP).} Unlike single-modal understanding, VLP aims to bridge vision and language by modeling their interactions. Generally, existing VLP methods can be categorized into two groups: (1) learning vision-language joint representations \cite{sun2019videobert,li2020oscar,li2020hero,DBLP:conf/icml/KimSK21,liu2021dimbert} and (2) learning visual representations from natural language supervision \cite{DBLP:conf/icml/RadfordKHRGASAM21,desai2021virtex,sariyildiz2020learning}. For the former, a deep cross-modal Transformer\cite{vaswani2017attention} is usually used to fuse multi-modal inputs and learn contextualized features. Among the latter, CLIP \cite{DBLP:conf/icml/RadfordKHRGASAM21} drew great attention because of its strong zero-shot transfer ability. Unlike the focused tasks in recent works \cite{luo2021clip4clip,shen2021much}, we analyze the effect of CLIP on video captioning.

\noindent\textbf{Multi-Task Learning (MTL).} The goal of MTL is to learn shared representations from multiple related tasks to improve the generalization performance of all tasks \cite{caruana1997multitask}. For video captioning, Pasunuru and Bansal \cite{pasunuru2017multi} proposed to train a video caption model with two directed-generation tasks, whose supervision signal was obtained from external datasets. By contrast, more attempts were made to construct auxiliary tasks by deriving additional supervision signal from the original annotations, \eg, predicting mined latent topics \cite{chen2017video} or extracted attributes \cite{yu2017end,li2019end,huang2020image} solely based on the input videos. The auxiliary task proposed in this paper instead takes either video content or textual descriptions as input.

\section{On INP \vs{} CLIP for Video Captioning}
\label{sec:AL}
This section aims to investigate the potential deficiencies of INP and explore the key to generating accurate descriptions. 
We organize this section as follows. We first briefly review video captioning in Sec.~\ref{sec:overview_of_vc}. Then, we introduce a Transformer baseline for video captioning in Sec.~\ref{sec:transformer_baseline}, where we will show how to integrate INP or CLIP models with the baseline. Finally, based on the experimental setup in Sec.~\ref{sec:AL_experimental_setup}, we present our analysis in Sec.~\ref{sec:AL_analysis}.

\begin{figure*}[t]
\centering
\includegraphics*[width = 0.99\linewidth]{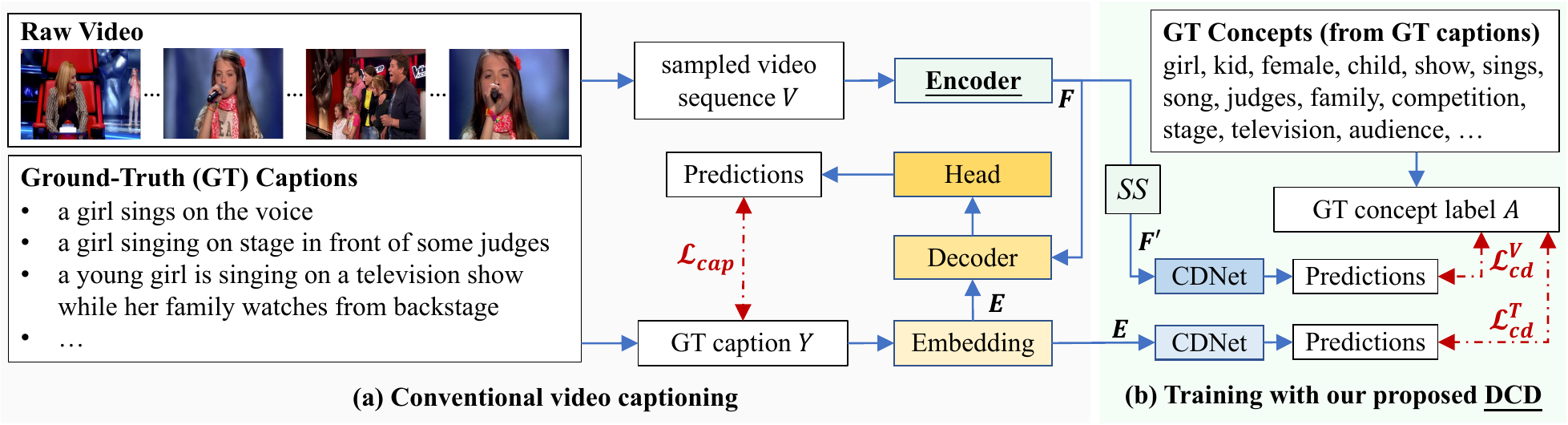}
\caption{Pipeline of video captioning at the training stage. In Sec.~\ref{sec:AL}, we will review (a) and focus on the encoder part of video caption models in which conventional methods usually use ImageNet pre-training models to encode the video content. In Sec.~\ref{sec:AR}, we will elaborate our proposed Dual Concept Detection (DCD) shown in (b), where ``SS'' denotes the sparse sampling.
}
\label{fig:pipeline}
\end{figure*}

\subsection{Overview of Video Captioning}
\label{sec:overview_of_vc}
As shown in Fig.~\ref{fig:pipeline} (a), conventional video captioning methods build upon the Encoder-Decoder framework. Formally, given a sequence of sampled video frames (or snippets) $V = \{v_1, v_2, \dots, v_N\}$ of length $N$, the caption encoder aims to encode $V$ into video features $\mathbf{F} \in \mathbb{R}^{d_h\times N}$:
\begin{equation}
\label{eq:encoding}
   \mathbf{F} = {\rm Encoder}(V).
\end{equation}
After the encoding stage, the decoding stage uses a typical autoregressive manner to generate a caption $Y = \{y_1, y_2, \dots, y_T\}$ of length $T$. Specifically at $t$-th time step, previously generated words $Y_{<t}$ are first embedded into low dimensional space to obtain text embeddings $\mathbf{E_{<t}} \in \mathbb{R}^{d_h\times t}$:
\begin{equation}
\label{eq:embedding}
\mathbf{E_{<t}} = {\rm Embedding}(Y_{<t}).
\end{equation}
Then given $\mathbf{E_{<t}}$ and video features $\mathbf{F}$, the rest decoding stage at $t$-th time step can be formulated as follows:
\begin{equation}
\label{eq:decoding}
\begin{split}
    h_t &= {\rm Decoder}(\mathbf{E_{<t}}, \mathbf{F}), \\
    p(y_{t}|Y_{<t}, V) &= {\rm softmax}({\rm Head}(h_t)),
\end{split}
\end{equation}
where the hidden state $h_t \in \mathbb{R}^{d_h}$ of the caption decoder is produced first and then fed to a classification head, followed by a softmax function to predict the probability distribution $p(y_{t}|Y_{<t}, V) \in \mathbb{R}^{|\mathcal{V}|}$ over the vocabulary $\mathcal{V}$. Finally, to train the video caption model, the cross entropy loss is used:
\begin{equation}
\label{eq:train_cap}
   \mathcal{L}_{cap} = - \sum_{t=1}^{T} \log p(y_{t}|Y_{<t}, V).
\end{equation}

\subsection{A Transformer Baseline for Video Captioning}
\label{sec:transformer_baseline}
Following the common practices \cite{zhang2021open,zhang2020object}, we implement the caption encoder as a pre-training backbone followed by a fully connected (FC) layer and a layer normalization (LN) layer \cite{ba2016layer}. Thus Eq.~\ref{eq:encoding} can be detailed as:
\begin{equation}
\label{eq:encoding_with_backbone}
   \mathbf{F} = {\rm Encoder}(V) = {\rm LN}({\rm FC}({\rm Backbone}(V))),
\end{equation}
where parameters in ${\rm LN}(\cdot)$ and ${\rm FC}(\cdot)$ are trainable while that of ${\rm Backbone(\cdot)}$ is fixed. We can replace the backbone in Eq.~\ref{eq:encoding_with_backbone} with either a INP model or a CLIP model to help encode the video content and then probe into the learned features $\mathbf{F}$ after training. When considering multi-modalities, we fuse features $\{\mathbf{F^{(m)}}\}_{m=1}^{M}$ of $M$ modalities along the sequence dimension to obtain $\mathbf{F} \in \mathbb{R}^{d_h\times MN}$. Details for the rest components are as follows. The embedding layer consists of a word embedding layer, a position embedding layer, and a LN layer, all of which are trainable. The caption decoder comprises a stack of Transformer decoder layers as in \cite{vaswani2017attention}. The classification head is implemented as a FC layer.

\subsection{Experimental Setup}
\label{sec:AL_experimental_setup}

\noindent\textbf{Dataset.} MSR-VTT\cite{xu2016msr} consists of 6,513, 497, and 2,990 video clips (20 English captions per clip) for training, validation, and testing, respectively.

\noindent\textbf{Metrics.} To measure caption quality, we use four common automatic metrics: CIDEr \cite{vedantam2015cider}, BLEU-4 \cite{papineni2002bleu}, METEOR \cite{banerjee2005meteor}, and ROUGE-L \cite{lin2004rouge}. All metrics are computed by Microsoft COCO Evaluation Server. We also report a summation of the front four metrics, named Meta-Sum. To quantify caption diversity, we follow \cite{yang2021non} to compute three metrics, including Novel (the percentage of captions that have not been seen in the training data), Unique (the percentage of captions that are unique among the generated captions), and Vocab (the number of words in the vocabulary that are adopted to generate captions). 

\noindent\textbf{Implementation Details.} Given a video, we uniformly sample $N=28$ frames (snippets) first and then use ResNet-101\cite{he2016deep} pre-trained on ImageNet, 3D ResNeXt-101\cite{hara2018can} pre-trained on Kinetics-400, and VGGish\cite{hershey2017cnn} pre-trained on AudioSet to respectively extract features of image, motion, and audio modalities. For a fair comparison, we also treat ResNet-101 as the backbone of our baseline model when using CLIP. More details are left to Sec.~\ref{sec:approach_experimental_setup}.

\subsection{Analysis}
\label{sec:AL_analysis}

\begin{table}[t]
    \centering
    \setlength\tabcolsep{3.2pt}
    \caption{Performance on MSR-VTT with ImageNet Pre-training (INP) or CLIP. We consider \textbf{I}mage, \textbf{Mo}tion and \textbf{A}udio modalities and report \textbf{C}IDEr, \textbf{B}LEU-\textbf{4}, \textbf{M}ETEOR, \textbf{R}OUGE-L, \textbf{M}eta-\textbf{S}um, \textbf{N}ovel, \textbf{U}nique, and \textbf{V}ocab metrics.}
    \label{tab:effect_of_clip} 
    \vspace{-5pt}
    \begin{tabular}{clccccc|ccc}  
    \toprule 
    \multirow{2}{*}[-0.5ex]{\#} &\multirow{2}{*}[-0.5ex]{Modality} &\multicolumn{5}{c}{Caption Quality} &\multicolumn{3}{c}{Caption Diversity}\cr \cmidrule{3-10}
    & &C &B4 &M &R &MS &N &U &V \cr
    \midrule
    1 &I$_{\rm INP}$           &41.5 &37.7 &26.9 &58.4 &164.5    &9.7   &17.2  &299 \cr
    2 &I$_{\rm INP}$ + Mo       &46.3 &39.9 &27.8 &59.8 &173.8    &13.5  &22.0  &357 \cr
    3 &I$_{\rm INP}$ + Mo + A   &50.0 &43.9 &29.2 &61.9 &185.0    &20.5  &30.4  &409 \cr
    \midrule
    4 &I$_{\rm CLIP}$           &52.8 &43.4 &29.6 &61.8 &187.6    &20.8  &30.7  &497 \cr
    5 &I$_{\rm CLIP}$ + Mo       &54.6 &44.6 &29.9 &62.7 &191.8    &24.0  &34.8  &\textbf{518} \cr
    6 &I$_{\rm CLIP}$ + Mo + A   &\textbf{55.2} &\textbf{47.1} &\textbf{30.4} &\textbf{63.6} &\textbf{196.3} &\textbf{24.2}  &\textbf{35.4} &492 \cr
    \bottomrule
    \end{tabular}
    \vspace{-10pt}
\end{table}

We conduct a series of analyses on MSR-VTT to answer the following questions.

\textbf{(1) Can CLIP improve video captioning compared with INP?} For a fair comparison, we treat ResNet-101 with either INP or CLIP as the backbone of our baseline model and keep other components the same. As we can see in Table~\ref{tab:effect_of_clip}, using CLIP brings significant improvements over INP on both caption quality and diversity under different combinations of modalities, \eg{}, compared with model 1, model 4 gains a relative improvement of 27.2\% at CIDEr and generates more than twice as many novel captions. Although taking more modalities as the model inputs gradually narrow the performance gap, a relative gain of 10.4\% at CIDEr for model 6 \vs{} model 3 is still overwhelming.

\begin{figure*}[t]
\centering
\includegraphics*[width = 0.8\linewidth]{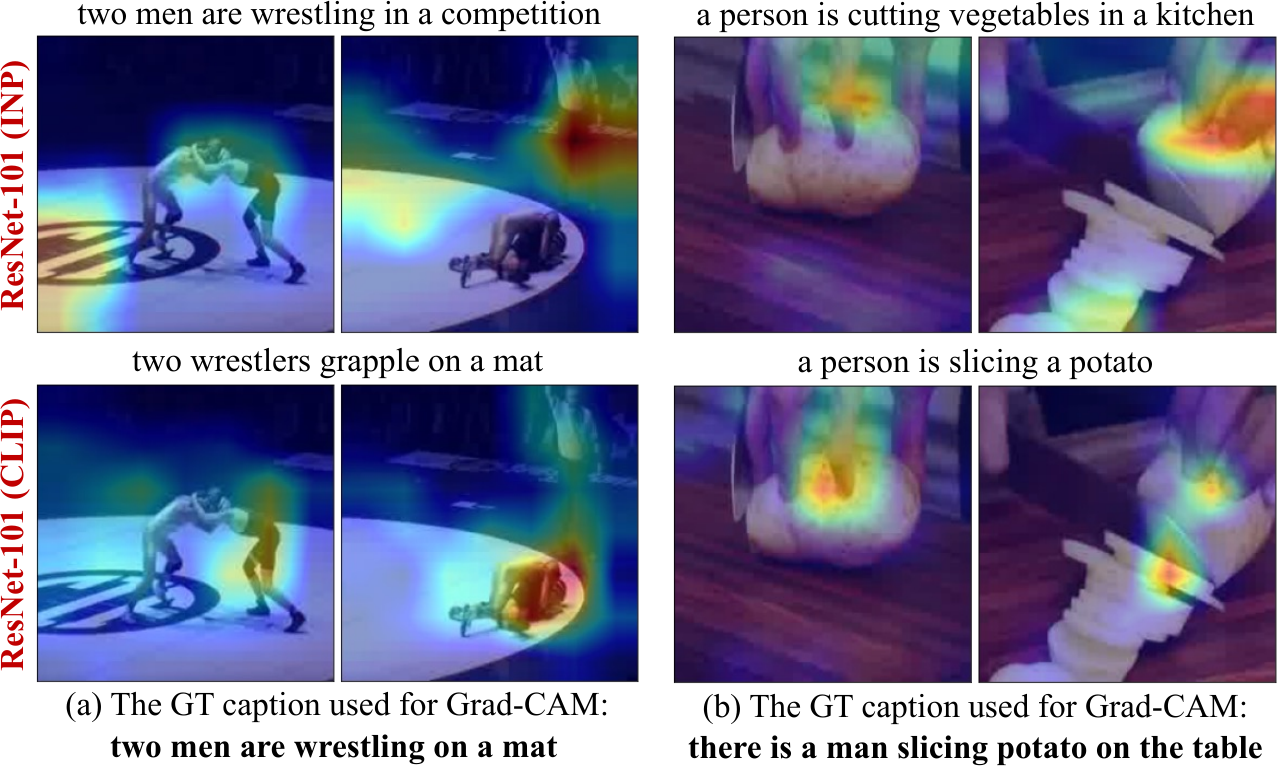}
\caption{Grad-CAM for the given ground-truth (GT) captions. In each case, the input of the video caption model is two manually selected keyframes, based on which the actual generated description is given above the subfigure.}
\label{fig:gradcam}
\end{figure*}

\begin{figure}[t]
    \centering  
    \includegraphics[width = 0.7\linewidth]{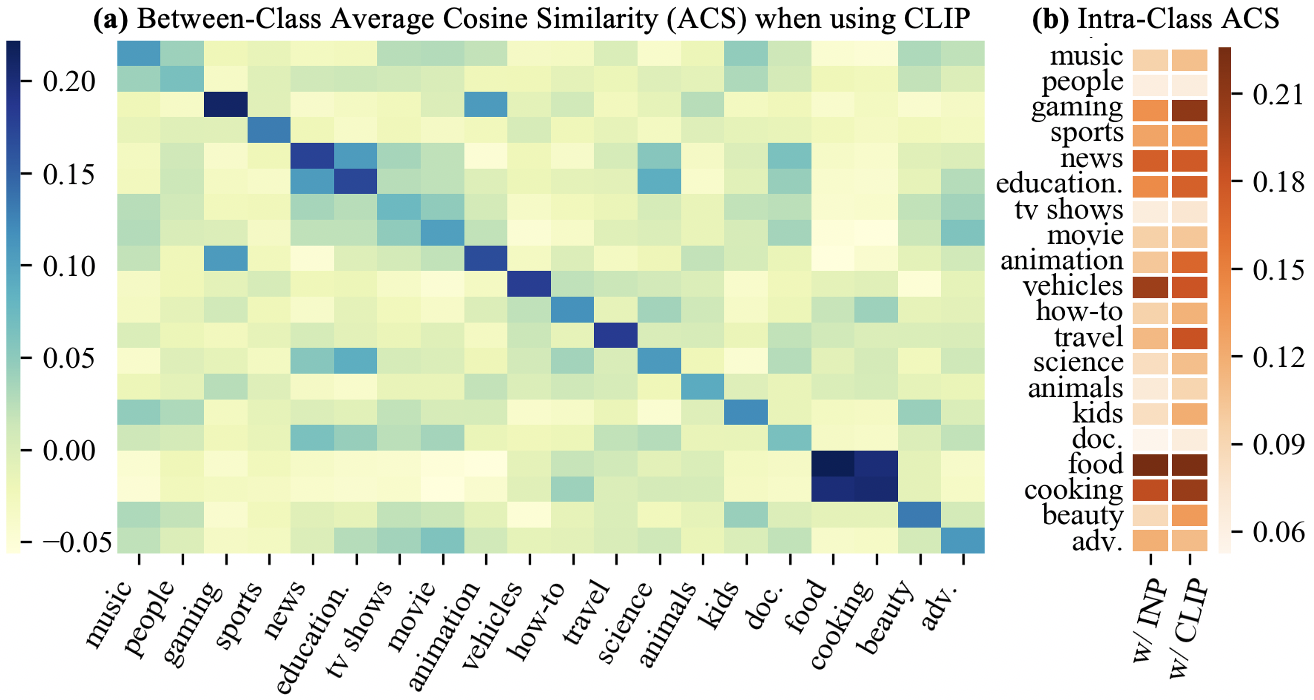}
    \caption{Quantitative analysis on the learned video features. In (b), the mean intra-class ACS over all categories is relatively improved 18.1\% after using CLIP.}
    \label{fig:impact_of_CLIP_on_ACS}
\end{figure}

\textbf{(2) What results in the large performance gap between INP and CLIP?} We answer this question from two aspects. The first aspect is to explore the decision-making process of models. We use Grad-CAM \cite{selvaraju2017grad}, a gradient-based localization technology, to figure out which part of the input videos is responsible for the models' outputs. As we can see the first row in Fig.~\ref{fig:gradcam}, using INP deviates the model's focus from the most critical regions for the given GT captions, \eg{}, the neglect of two wrestlers that grapple on a mat in (a) and the wrong attention towards fingers in (b). By contrast, the second row in Fig.~\ref{fig:gradcam} shows that using CLIP enables the model to be more aware of where the concepts are. This is conducive to accurate captioning, \eg{}, ``vegetables'' \vs{} ``potato'' in (b).

\begin{wrapfigure}[12]{r}{0.62\linewidth}
    \vspace{-20pt}
    \centering  
    \includegraphics[width = 1\linewidth]{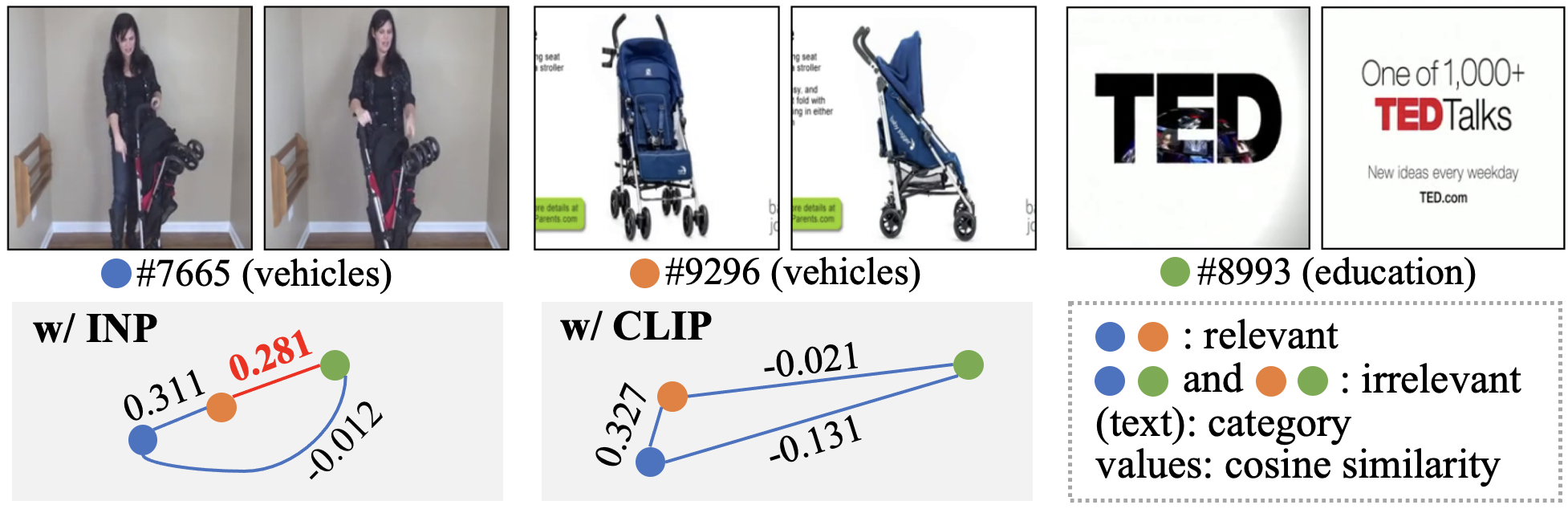}
    \caption{An example on the learned video features. Using CLIP enables the video caption model to encode more semantics of the concepts ``stroller'' and ``TED'' rather than the background.}
    \label{fig:INP_vs_CLIP_example}
\end{wrapfigure}

The second aspect is to analyze the learned video features. We first obtain mean pooled video features for videos in the testing set, followed by the z-score normalization. Then we calculate between- or intra-class Average Cosine Similarity (ACS) based on the ground-truth category tags of MSR-VTT. As shown in Fig.~\ref{fig:impact_of_CLIP_on_ACS} (a), the video caption model can generally distinguish features from different categories when using CLIP (so does the model using INP). In (b), we can clearly observe the difference between INP and CLIP, \ie{}, using CLIP can learn more similar intra-class video features on almost all categories, indicating that using CLIP can better capture the topic-related characteristics of video content. Moreover, we give a qualitative example in Fig.~\ref{fig:INP_vs_CLIP_example}, where the first two videos share similar concepts ``stroller'' and the last two videos have a similar white background. We can see that a high similarity is wrongly assigned to \#9296 and \#8993 when using INP while using CLIP makes the model more robust against irrelevant background information and more sensitive to the concepts. 

\textbf{Main conclusions:} (1) INP does not work well in video captioning, which manifests as the unsatisfying captioning performance in Table~\ref{tab:effect_of_clip}, the overlook of concepts' semantics in Fig.~\ref{fig:gradcam}, and the vulnerability to irrelevant background interference in Fig.~\ref{fig:INP_vs_CLIP_example}. The deficiencies of INP can be attributed to the domain gap between ImageNet data and video captioning data. (2) Using CLIP helps describe the accurate videos' details, which in turn promotes caption diversity. The quantitative and visualized results suggest that concept-aware representation learning is the key to prompting accurate captioning.

\section{Approach}
\label{sec:AR}
Based on above findings, we propose to further strengthen concept-aware representation learning of a video caption model via Dual Concept Detection (DCD). Next, we fisrtly elaborate the formulation of concept detection in Sec.~\ref{sec:approach_formulation}, followed by our proposed DCD in Sec.~\ref{sec:approach_DCD}. 
Then based on the experimental setup in Sec.~\ref{sec:approach_experimental_setup}, we compare our approach with state-of-the-art methods in Sec.~\ref{sec:comparison_with_SOTA} and carry out ablation studies to verify the effectiveness of DCD in Sec.~\ref{sec:AR_analysis}.

\subsection{Formulation of Concept Detection}
\label{sec:approach_formulation}
\textbf{Supervision Signal.} Given training captions, we filter out stop words and keep the most frequent $K$ words as the targets. Then we can obtain a multi-hot concept label $A=\{a_1, a_2, \dots, a_K\}$ for each video, where $a_k = 1$ indicates that the $k^{th}$ concept exists in ground-truth captions of the video.

\noindent\textbf{Feed-Forward Pass.} Let denote $\mathbf{X} \in \mathbb{R}^{d_h\times L}$ as the features used for concept detection, $d_h$ the hidden dimension of the model, and $L$ the number of instances. We feed $\mathbf{X}$ into an concept detection network (CDNet) followed by a sigmoid function to obtain a raw probability matrix $\mathbf{P_{raw}} \in \mathbb{R}^{K\times L}$:
\begin{equation}
\label{eq:ar_raw}
   \mathbf{P_{raw}} = {\rm sigmoid}({\rm CDNet}(\mathbf{X})),
\end{equation}
where $P_{raw}^{k, l}$ denotes the possibility that the $l^{th}$ instance contains the $k^{th}$ concept. Then for the $k^{th}$ concept, we adopt multiple instance learning to merge probability values of all $L$ instances and obtain the final probability $p_k$:

\begin{equation}
\label{eq:noisy_or}
    \forall k \in [1, K],\quad p_k = 1 - \prod^L_{l=1}(1 - P_{raw}^{k, l}).
\end{equation}

\noindent\textbf{Training Objective.} We here use the binary cross entropy loss:
\begin{equation}
\label{eq:ar_bce}
    \mathcal{L}_{bce} = - \frac{1}{K_{pos}} \sum^{K}_{k=1} (a_k\log(p_k)+(1-a_k)\log(1-p_k)),
\end{equation}
where the loss for each sample is normalized by the number of positive concepts $K_{pos}$. In practice, we find that when solely using $\mathcal{L}_{bce}$, $p_k$ in Eq.~\ref{eq:noisy_or} is prone to be $1$ for all concepts at the beginning of training, leading to the so-called gradient vanishing problem. Thus, we propose to add an extra regularization term $\mathcal{L}_{reg}$:

\begin{equation}
\label{eq:ar_reg}
    \mathcal{L}_{reg} = \max({\rm mean}(\mathbf{P_{raw}}) - K_{pos}/K, 0).
\end{equation}
With the proposed $\mathcal{L}_{reg}$, concept detection is forced to be conservative, and the training procedure becomes stable. The final training objective $\mathcal{L}_{cd}$ is:
\begin{equation}
\label{eq:ap}
    \mathcal{L}_{cd} = \mathcal{L}_{bce} + \mathcal{L}_{reg}.
\end{equation}

\subsection{Dual Concept Detection (DCD)}
\label{sec:approach_DCD}
DCD is an auxiliary task requiring the video caption model to infer relevant concepts based on partial information from videos or text descriptions. Thus, DCD carries out the following two types of detection.

\noindent\textbf{Video-based Concept Detection} is expected to predict the concepts of a video (\ie{}, $A$) given sparsely sampled video features $\mathbf{F'}$. Specifically, given video features $\mathbf{F} \in \mathbb{R}^{d_h\times N}$ after encoding (Eq.~\ref{eq:encoding_with_backbone}), we introduce a sampling ratio $r \in (0, 1]$ to obtain $\mathbf{F'} \in \mathbb{R}^{d_h\times N'}$, where $N' = \lceil N\cdot r \rceil$. In implementation, $r$ is set to be distributed uniformly during training. Finally, we replace $\mathbf{X}$ in Eq.~\ref{eq:ar_raw} with $\mathbf{F'}$ and get a resulting training objective $\mathcal{L}^{V}_{cd}$.

\noindent\textbf{Text-based Concept Detection} aims to predict the concepts of a video solely based on a corresponding ground-truth (GT) caption that contains limited observable concepts. Formally, given a GT caption $Y$ of length $T$, we first obtain its embedding representation $\mathbf{E} \in \mathbb{R}^{d_h\times T}$:
\begin{equation}
\label{eq:AR_T}
\forall y_t \in Y, e_t \in \mathbf{E},\quad e_t = {\rm LN}(\mathbf{W}^{w}_{[y_t]} + \mathbf{W}^{p}_{[t]}),
\end{equation}
where $\mathbf{W}^{w} \in \mathbb{R}^{d_h\times |\mathcal{V}|}$ is word embeddings of the vocabulary $\mathcal{V}$ and $\mathbf{W}^{w}_{[y_t]}$ yields a embedding vector indexed by $y_t$; $\mathbf{W}^{p} \in \mathbb{R}^{d_h\times T_{max}}$ is positional embeddings while $T_{max}$ is a pre-defined maximum sequence length. Then, we replace $\mathbf{X}$ in Eq.~\ref{eq:ar_raw} with $\mathbf{E}$ and get a resulting training objective $\mathcal{L}^{T}_{cd}$. 

\noindent\textbf{Overall Training Objective.} We jointly optimize $\mathcal{L}_{cap}$ (Eq.~\ref{eq:train_cap}) for caption generation and $\mathcal{L}^{V}_{cd}$ and $\mathcal{L}^{T}_{cd}$ for DCD:
\begin{equation}
\label{eq:overall_objective}
    \mathcal{L} = \mathcal{L}_{cap} + \mathcal{L}^{V}_{cd} + \mathcal{L}^{T}_{cd}.
\end{equation}

\subsection{Experimental Setup}
\label{sec:approach_experimental_setup}
\noindent\textbf{Datasets and Metrics.} Apart from MSR-VTT mentioned in Sec.~\ref{sec:AL_experimental_setup}, we also experiment on VATEX \cite{wang2019vatex}, which contains 41,269 video clips (10 English descriptions per clip) and is divided into 25,991 training, 3,000 validation, and 6,000 testing. We use all metrics introduced in Sec.~\ref{sec:AL_experimental_setup}.

\noindent\textbf{Implementation Details.} For concept detection, we set the number of concepts $K$ to 500 and implement CDNet in Eq.~\ref{eq:ar_raw} as a fully connected layer. For our baseline model, we set the number of Transformer decoder layers to 1 and set the hidden dimension $d_h$ to 512 for MSR-VTT whereas 1,024 for VATEX. We set the maximum length of sentences $T_{max}$ to 30, and train batches of 64 video-sentence pairs using ADAM \cite{kingma2014adam} with an initial learning rate of 5e-4 and L2 weight decay of 0.001. Beam search with a beam size of 5 is used for inference. We by default use all three modalities of videos based on Table~\ref{tab:effect_of_clip} and take CLIP's ViT-B/32 encoder as the image backbone based on Table~\ref{tab:clip_variants}.

\begin{table}[t]
    \centering
    \setlength\tabcolsep{3.2pt}
    \caption{Meta-Sum scores of different visual encoders of CLIP.}
    \vspace{-5pt}
    \label{tab:clip_variants}  
    \begin{tabular}{ccccc}  
    \toprule
    &ResNet-50 &ResNet-101 &ResNet-50x4 &ViT-B/32 \cr 
    \midrule
    MSR-VTT &193.9 &196.3 &198.2 &\textbf{199.2}\cr
    VATEX &168.3 &171.9 &\textbf{174.8} &174.4 \cr
    \bottomrule
    \end{tabular}
\end{table}

\begin{table*}[t]
\begin{floatrow}
\capbtabbox{
\vspace{-5pt}
\fontsize{8}{10}\selectfont  
\begin{tabular}{cccccc}
    \toprule
    Method &Year &C &B4 &M &R \cr 
    \midrule
    ORG-TRL\cite{zhang2020object} &2020 &50.9 &43.6 &28.8 &62.1 \cr
    MGCMP\cite{chen2021motion} &2021 &51.4 &41.7 &28.9 &62.1\cr
    NACF\cite{yang2021non} &2021 &51.4 &42.0 &28.7 &- \cr
    APML\cite{lin2021augmented} &2021 &52.2 &43.8 &30.3 &63.6\cr
    OpenBook\cite{zhang2021open} &2021 &52.9 &42.8 &29.3 &61.7 \cr
    ARB-ACL\cite{li2022adaptive} &2022 &51.3 &42.6 &28.9 &61.5 \cr 
    \midrule
    CLIP-Base &Ours &57.0 &47.3 &30.9 &64.0 \cr
    CLIP-DCD &Ours &\textbf{58.7} &\textbf{48.2} &\textbf{31.3} &\textbf{64.8}\cr
    \bottomrule
\end{tabular}
}{
 \caption{Comparison on MSR-VTT. }
 \label{tab:sota_msrvtt}
}

\capbtabbox{
\vspace{-5pt}
\fontsize{8}{10}\selectfont  
\begin{tabular}{cccccc}
    \toprule
    Method &Year &C &B4 &M &R \cr 
    \midrule
    VATEX\cite{wang2019vatex} &2019 &45.1 &28.4 &21.7 &47.0 \cr
    ORG-TRL\cite{zhang2020object} &2020 &49.7 &32.1 &22.2 &48.9 \cr
    NSA\cite{guo2020normalized} &2020 &57.1 &31.0 &22.7 &49.0 \cr
    OpenBook\cite{zhang2021open} &2021 &57.5 &33.9 &23.7 &50.2 \cr
    MGCMP\cite{chen2021motion} &2021 &57.6 &34.2 &23.5 &50.3\cr
    \midrule
    CLIP-Base &Ours &60.9 &\textbf{36.8} &24.8 &51.9\cr
    CLIP-DCD &Ours &\textbf{62.4} &\textbf{36.8} &\textbf{25.1} &\textbf{52.2} \cr
    \bottomrule
\end{tabular}
}{
 \caption{Comparison on VATEX.}
 \label{tab:sota_vatex}
}
\vspace{-10pt}
\end{floatrow}
\end{table*}

\subsection{Comparison with State-of-the-Art Methods}
\label{sec:comparison_with_SOTA}
We compare our approach with state-of-the-art methods (SOTAs) on MSR-VTT and VATEX in Table ~\ref{tab:sota_msrvtt} and \ref{tab:sota_vatex}. We can observe that CLIP-Base already surpasses SOTAs by a large margin due to the better learning of concept-aware representations mentioned in Sec.~\ref{sec:AL_analysis}. But we note that the relative improvement of CLIP-Base against SOTAs on VATEX is relatively small than that of MSR-VTT. This is probably because videos in VATEX are originated from the action recognition dataset, making the motion cues more important than the static appearance. Compared with CLIP-Base, the model trained with our proposed DCD (\ie{}, CLIP-DCD) performs better on both datasets, especially the CIDEr metric, \eg{}, a relative improvement of 3.0\% on MSR-VTT.

\begin{table*}[b]
    \centering
    \fontsize{8.5}{10}\selectfont  
    \setlength\tabcolsep{4pt}
    \caption{Impact of using different training objectives on MSR-VTT.}
    \label{tab:ablation_study_main} 
    \vspace{-5pt}
    \begin{tabular}{cccc|ccccc|ccc}  
    \toprule 
    \multirow{2}{*}[-0.5ex]{Model}
    &\multicolumn{3}{c}{Objective} 
    &\multicolumn{5}{c}{Caption Quality}
    &\multicolumn{3}{c}{Caption Diversity}
    \cr\cmidrule{2-12}
    &$\mathcal{L}_{cap}$ &$\mathcal{L}^{V}_{cd}$ & $\mathcal{L}^{T}_{cd}$
    &C &B4 &M &R &MS
    &N &U &V
    \cr\midrule
    
    CLIP-Base &\checkmark & & 
    &57.0 &47.3 &30.9 &64.0 &199.2 
    &28.0 &40.0 &520\cr
    
    CLIP-VCD &\checkmark &\checkmark &
    &58.4 &48.1 &\textbf{31.3} &\textbf{64.8} &202.6 
    &31.5 &43.2 &565\cr
    
    CLIP-TCD &\checkmark & &\checkmark 
    &58.0 &47.6 &31.1 &\textbf{64.8} &201.5
    &\textbf{31.7} &\textbf{43.9} &\textbf{581}\cr
    
    CLIP-DCD &\checkmark &\checkmark &\checkmark 
    &\textbf{58.7} &\textbf{48.2} &\textbf{31.3} &\textbf{64.8} &\textbf{203.0}
    &30.7 &43.2 &569\cr

    \bottomrule
    \end{tabular}
\end{table*}

\subsection{Ablation Study}
\label{sec:AR_analysis}
We here experiment on MSR-VTT to delve deeper into DCD's design. 

\begin{figure}[t]
    \centering
    \includegraphics*[width = 0.9\linewidth]{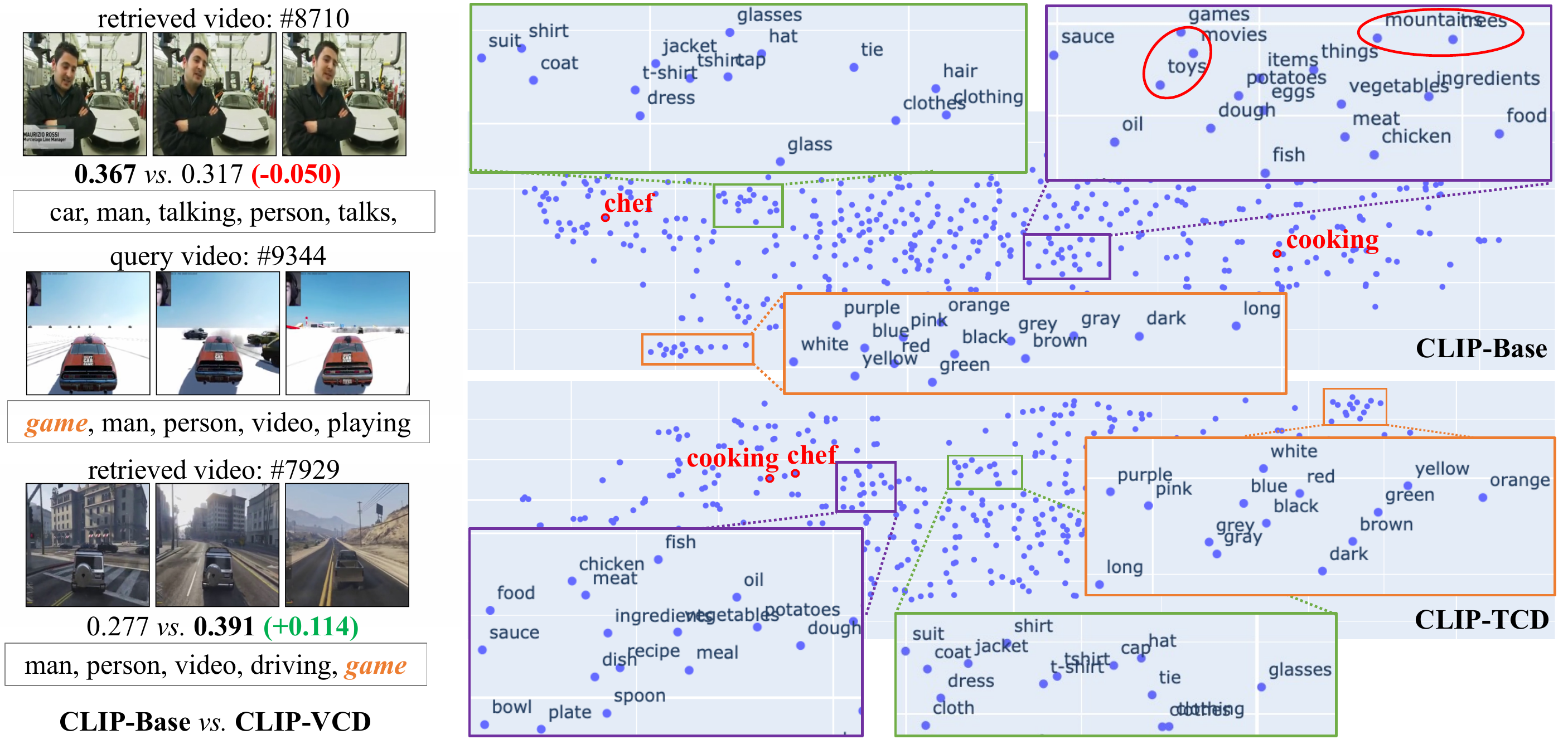}
    \caption{(Left): Retrieval comparison for CLIP-Base \vs{} CLIP-VCD. Values are cosine similarities and words are CLIP-VCD's top 5 detected concepts. (Right): Analysis on the learned concept embeddings via t-SNE visualization.}
    \label{fig:VCD_DCD}
\end{figure}

\noindent\textbf{Impact of Video-based Concept Detection (VCD).} Results in Table~\ref{tab:ablation_study_main} show that using VCD can improve all metrics (CLIP-VCD \vs{} CLIP-Base). We further give an example in Fig.~\ref{fig:VCD_DCD} (left) to illustrate the difference between CLIP-Base and CLIP-VCD. As we can see, while similar retrieval results can be obtained for the given query video, two models have different preferences, \eg{}, CLIP-VCD prefers videos with a similar style to the query video. This can be explained by the detected concepts of CLIP-VCD, where the concept ``game'' is detected in video \#9344 and \#7929.

\noindent\textbf{Impact of Text-based Concept Detection (TCD).} 
Similar to VCD, using TCD can also improve the captioning performance, \ie{}, CLIP-TCD \vs{} CLIP-Base in Table~\ref{tab:ablation_study_main}. To gain more insights into TCD, we extract $K=500$ concept embeddings from word embeddings (\ie{}, $\mathbf{W}^{w}$ in Eq.~\ref{eq:AR_T}) and visualize them via t-SNE \cite{van2008visualizing}. Based on Fig.~\ref{fig:VCD_DCD} (right), we have the following two observations. 
(1) For CLIP-Base with only the caption generation objective, concepts of the same type are generally clustered together. But some noises may exist in those potential clusters, \eg{}, concepts like ``toys'' and ``games'' are contained in the cluster about food and tableware (the purple bounding box). 
(2) After using TCD, the co-occurrence relations between concepts affect the learned representations, \eg{}, ``cooking'' and ``chef'', two concepts that often co-occur with food and tableware, become closer to the purple bounding box in (b).

\begin{figure*}[t]
\centering
\includegraphics*[width = 0.95\linewidth]{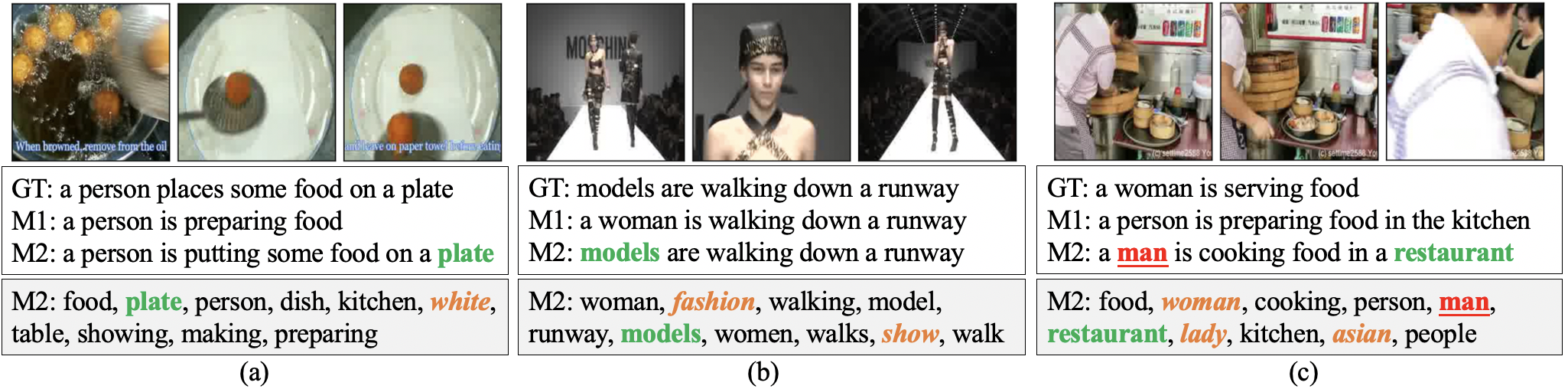}
\caption{Captioning results of the CLIP-Base (M1) and CLIP-DCD (M2) models, where we emphasize \textcolor{darkgreen}{\textbf{accurate keywords}} and \textcolor{red}{\textbf{\underline{errors}}}. Among M2's video-based detected concepts, we highlight \textcolor{orange}{\textbf{\textit{meaningful concepts}}} yet to be utilized.
} 
\label{fig:qualitative_examples}
\end{figure*}

\noindent\textbf{Qualitative results.} Fig.~\ref{fig:qualitative_examples} presents three examples to illustrate the difference between CLIP-Base's and CLIP-DCD's generated captions. We can see that while CLIP-Base describes the main event of videos accurately, CLIP-DCD can capture more details, \ie{}, ``plate'' in (a), ``models'' in (b), and ``restaurant'' in (c). We note that CLIP-DCD mistakes the woman in (c) as a man. One reason might be that it is difficult to recognize the woman from behind. Another reason might be that the language model is biased by the imbalanced training data, where the word ``man'' is more frequent than ``woman''. However, given the video-based detected concepts (\ie{}, VCD's results) of CLIP-DCD below each example, we can see that there is a great potential to use the detected concepts to polish the captioning results of CLIP-DCD, \eg{}, ``white plate'' in (a), ``fashion show'' in (b), and ``Asian woman (lady)'' in (c).

\section{Conclusions}
This paper carries out an empirical study on INP \vs{} CLIP, where we reveal the potential deficiencies of INP and conclude that concept-aware representation learning contributes significantly to accurate video captioning. Motivated by these findings, we devise an auxiliary task named Dual Concept Detection to inject concept-related knowledge into video caption models during training. Experiments show that our approach enables better learning of concept-aware video and word representations. 

\section*{Acknowledgements}
\footnotesize This paper was partially supported by NSFC (No: 62176008) and Shenzhen Science \& Technology Research Program (No:GXWD20201231165807007-20200814115301001).

%
%
%

\bibliographystyle{splncs04}
\bibliography{egbib}
\end{document}